\documentclass[11pt]{article}

\usepackage[margin=1.1in]{geometry}
\usepackage[T1]{fontenc}
\usepackage[utf8]{inputenc}
\usepackage{microtype}
\usepackage{booktabs}
\usepackage{array}
\usepackage{tikz}
\usetikzlibrary{positioning, fit, arrows.meta}
\usepackage[numbers,sort&compress]{natbib}
\usepackage{xcolor}
\usepackage{enumitem}
\setlist{itemsep=2pt, topsep=3pt, parsep=0pt}
\usepackage{xurl}
\usepackage[hidelinks]{hyperref}
\usepackage{graphicx}

\newcommand{\code}[1]{\texttt{#1}}
\title{Choir: An Open Protocol for Distributed Multi-Agent Autoformalization}

\author{
  Yidi Qi\\
  Harvard University\\
  \texttt{yidi\_qi@seas.harvard.edu}
  \and
  Melanie Weber\\
  Harvard University\\
  \texttt{mweber@seas.harvard.edu}
}

\date{}

\begin{document}
\maketitle

\begin{center}
  \includegraphics[width=0.18\textwidth]{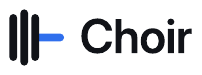}
\end{center}
\vspace{.5em}

\begin{abstract}
AI agents can now formalize entire textbooks and major theorems in proof assistants such as Lean, but current efforts are typically centralized: a single team runs all agents and bears the full computational cost. We introduce \emph{Choir}, an open protocol for distributed formalization. Choir decomposes a project into tasks that can be completed by independent contributors, each running their own agent with their own LLM subscription, while coordinating entirely through the project’s GitHub repository. To support open participation, every contribution is checked by a deterministic gate before merge. Choir supports Lean~4, Isabelle, and Rocq, and is open source and modular, allowing projects to replace individual components or extend the protocol.
\end{abstract}

\section{Introduction}
\label{sec:intro}

Proof assistants such as Lean~4~\citep{demoura2021lean4},
Isabelle~\citep{nipkow2002isabelle}, and Rocq~\citep{rocq2025} check every
step of a proof against a small trusted kernel, so a formalized result can be
relied on without re-reading its proof. Communities maintain shared
libraries on top of them, such as the Lean libraries Mathlib~\citep{mathlib2020},
Physlib~\citep{physlib2026,toobysmith2024heplean} and
CSLib~\citep{barrett2026cslib} for mathematics, physics and computer science,
which new formalizations build on. Until recently this work was done almost
entirely by hand, and building these libraries took many contributors years.

AI agents have changed the pace. In the past year they have formalized whole
textbooks. A Meta FAIR team formalized a graduate algebraic combinatorics
textbook into roughly 130{,}000 lines of Lean in one week, using about
30{,}000 agent runs working in parallel at an estimated cost of \$100K with prompt
caching and \$430K without~\citep{gloeckle2026textbook}. The same group has
since built Atlas, a library formalized from 26
textbooks~\citep{rammal2026scale}. This effort is part of a broader shift towards large-scale, agent-driven formalization across mathematics~\citep{wang2026m2f,meek2026numerical,bryant2026munkres,urban2026megalodon}.
Agents have also completed major research-level formalizations. Math~Inc.'s
Gauss finished the formal proofs of sphere packing in dimensions 8 and 24, the
first building on a human-led
project~\citep{hariharan2026sphere,mathinc2026sphere}. Most
recently, a team of Claude agents at Anthropic completed a formal proof of
Fermat's Last Theorem in Lean in under two weeks~\citep{anthropic2026flt}.
Most of these systems are built for a single team that runs every agent on its
own budget, and several operate at a scale out of reach for a typical academic
group.

The open mathematical community has what such efforts need: expertise, and a
large amount of agent capacity, already paid for and spread across individual
LLM subscriptions. If that capacity could be pooled, large-scale formalization
would no longer depend on one well-funded team. Concurrent projects such as
\emph{TauCeti}~\citep{tauceti2026} and \emph{Prove2Me}~\citep{prove2me2026} explore this
idea (Section~\ref{sec:related}).

We present \emph{Choir}, an open protocol for distributed formalization.
Choir decomposes a project into tasks that can be completed by independent contributors,
each running their own agent with their own LLM subscription, while coordinating entirely
through the project’s GitHub repository. The project's overseer runs an
orchestrator agent that plans the project, writes the theorem statements, and
publishes them as tasks in a GitHub repository. Contributors claim tasks, prove
them on their own machines, and submit pull requests. Since contributors may be
strangers running unknown models, a deterministic verification gate checks
each pull request, including that the published statement is unchanged, before
the orchestrator reviews and merges it. Choir currently supports 
Lean~4, Isabelle, and Rocq. It is completely open source and designed to be modular:
a project can swap in its own planner or orchestrator, and contributors can use
any agent. Our hope is that others will be able to build on Choir, or combine
it with their own workflows.

\paragraph{Contributions.} This paper makes three technical contributions:
\begin{enumerate}[label=(\arabic*), leftmargin=2.2em]
\item \textbf{A task distribution and management system}
  (Section~\ref{sec:tasks}) that distributes a project's formalization across
  many contributors' agents through its own GitHub repository, and lets the
  proof tree grow from the decompositions they send back.
\item \textbf{A deterministic verification gate} (Section~\ref{sec:gate})
  that re-checks every pull request from source mechanically.
\item \textbf{A history visualizer} (Section~\ref{sec:viz}) that replays how a
  project's proof tree grew and which contributor proved or decomposed each
  node, reconstructed from its git log, issues, and pull requests alone.
\end{enumerate}

We tested Choir by formalizing Yufei Zhao's lecture notes \emph{Probabilistic
Methods in Combinatorics}~\citep{zhao2024probmethods} in Lean~4. The same notes
are among the textbooks in Meta FAIR's Atlas~\citep{rammal2026scale}, which
makes a direct comparison possible. The result is at
\url{https://github.com/yidiq7/ProbMethodCombinatorics}, and the code of Choir
is available under the Apache-2.0 license at
\url{https://github.com/Weber-GeoML/Choir}.

\section{Task distribution and management}
\label{sec:tasks}

\begin{figure}[t]
\centering
\includegraphics[width=\textwidth]{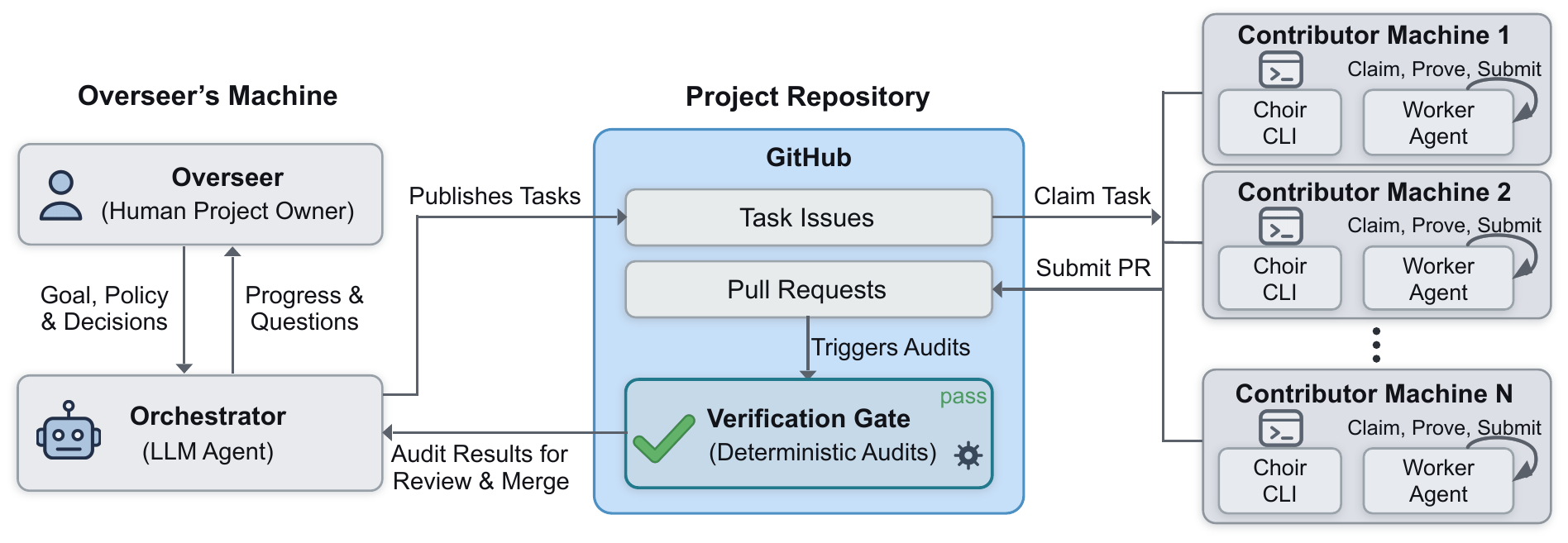}
\caption{The orchestrator publishes tasks as GitHub issues, and workers on
contributors' own machines claim them and submit pull requests. The gate
audits each pull request before the orchestrator reviews and merges it.}
\label{fig:arch}
\end{figure}

Choir splits a project across four parties (Table~\ref{tab:roles},
Figure~\ref{fig:arch}): a human overseer, two kinds of LLM agents, and the
repository itself. This section follows one unit of work from the
orchestrator's plan to a merged pull request, naming the commands each step
runs.

\begin{table}[h]
\centering
\footnotesize
\begin{tabular}{@{}>{\raggedright\arraybackslash}p{1.9cm}>{\raggedright\arraybackslash}p{3.2cm}>{\raggedright\arraybackslash}p{10.0cm}@{}}
\toprule
\textbf{Role} & \textbf{Runs where} & \textbf{Does} \\
\midrule
Overseer & --- (human) & Owns the project. Sets the goal, the audit policy, and
how far the orchestrator may act alone, and rules on new axioms. \\
\addlinespace[2pt]
Orchestrator & overseer's machine, LLM account, and GitHub auth &
Plans and decomposes the goal, \emph{authors the statements}, publishes typed
tasks, reviews and merges pull requests, replans in a loop. \\
\addlinespace[2pt]
Workers & contributors' machines and accounts & Claim a task, prove it with the
contributor's own agent, submit. Hold \emph{no} repository write access: a
claim is a comment, a submission is a pull request from the contributor's own
fork. \\
\addlinespace[2pt]
Gate & GitHub Actions in the project repo & Deterministic audits and intake
validation. \\
\bottomrule
\end{tabular}
\caption{The four roles.}
\label{tab:roles}
\end{table}

Choir manages the file edits and the version control
between multiple agents through GitHub, which already
supplies the tools a coordination platform needs: identity, event triggers, access control, public audit, etc. 
The commands in Table~\ref{tab:commands} bound what an agent may do.
 They are implemented as
Python wrappers over GitHub's \code{gh} CLI.
The orchestrator's procedures are written in a playbook document, so any
agent that can read the playbook and run these commands can serve as the
orchestrator.

\begin{table}[h]
\centering
\footnotesize
\begin{tabular}{@{}>{\raggedright\arraybackslash}p{1.9cm}>{\raggedright\arraybackslash}p{5.3cm}>{\raggedright\arraybackslash}p{7.9cm}@{}}
\toprule
\textbf{Party} & \textbf{Command} & \textbf{What it does} \\
\midrule
Orchestrator & \code{choir orch poll} & Blocks until a task or a pull request
changes, then exits with a JSON summary of what changed. The loop's clock. \\
\addlinespace[2pt]
 & \code{choir orch create-task} & Publishes a task issue from a plan node \\
\addlinespace[2pt]
 & \code{choir orch prs} & Open pull requests, with the verification gate's results \\
\addlinespace[2pt]
 & \code{choir orch merge\,$\mid$\,close} & Merges a pull request, or closes it unmerged.
\code{merge} refuses unless the gate passes.\\
\addlinespace[2pt]
 & \code{choir orch comment} & Posts a comment on a task issue or a pull
request \\
\addlinespace[2pt]
 & \code{choir orch sync-leases} & Reads the comments from workers and updates task labels\\
\addlinespace[2pt]
 & \code{choir orch sync-graph} & Rederives the roadmap graph's dependency
edges from the built project \\
\addlinespace[2pt]
 & \code{choir orch inventory scan} & Every axiom and placeholder in the
project with locations \\
\addlinespace[2pt]
 & \code{choir orch metrics struggle} & Per-task attempt history, recomputed
from GitHub \\
\addlinespace[2pt]
 & \code{choir orch salvage} & Opens a follow-up seeded from a failed pull
request \\
\addlinespace[2pt]
 & \code{choir orch viz} & Renders the project's history
(Section~\ref{sec:viz}) \\
\midrule
Worker & \code{choir worker list} & Claimable tasks \\
\addlinespace[2pt]
 & \code{choir worker claim} & Takes the lease and, if won, builds the
workspace \\
\addlinespace[2pt]
 & \code{choir worker heartbeat\,$\mid$\,release} & Refreshes a lease; hands one
back \\
\addlinespace[2pt]
 & \code{choir worker submit} & Pushes the branch and opens the pull request \\
\bottomrule
\end{tabular}
\caption{Choir's most-used commands.}
\label{tab:commands}
\end{table}

\paragraph{Plan.} The orchestrator opens a project by writing a roadmap into
the repository: the goal, the route it intends to take, its own reasoning log,
and a dependency graph as JSON. A node is publishable once it and every
dependency it names have been formally stated.
Decomposition of theorems at this stage is deliberately shallow to reduce the burden on
the orchestrator. The orchestrator states only the skeleton that carries the mathematical structure
and leaves further decomposition to the workers.

The planner is modular: the overseer can plug in their own planner or human-written files,
and use Choir only for task management, as long as the roadmap follows the same format.

\paragraph{Publish.} \code{choir orch create-task} turns a ready node into a
GitHub issue whose body opens with a YAML record.
Each task is a statement whose proof is a placeholder, labeled by the
orchestrator with a difficulty and a priority level. The overseer may cap the
difficulty a given worker will take based on which model they use, both saving costs 
and keeping weaker models off work they are unlikely to finish.

\paragraph{Claim.} A contributor working from another account holds no write
access to the project repository, so a worker claims a task by commenting on
its issue, as in human collaborations~\citep{buzzard2025flt,etp2025}. When
multiple workers reach for the same task, the earliest comment wins.
A claim that falls silent for a day expires and the task
returns to the pool. On a won claim the same call builds the workspace, which
is the project at the task's pinned commit on a task branch.

\paragraph{Prove and submit.} Proving happens on the contributor's own machine
under the contributor's own account. Choir ships no proving agent: a
contributor can use any agent or workflow, as long as it follows Choir's task
format and submission rules. \code{choir worker
heartbeat} keeps a longer lease alive by editing the worker's own comment. When the
proof is done, \code{choir worker submit} pushes the branch and opens the pull
request. \code{choir worker release} hands a task back cleanly, and a scheduled
sweep returns long-quiet claims to the pool.

\paragraph{Decompose.} Sometimes a published task is too hard. A worker
who finds that the mathematics does not fit one pull request may decompose it
further: state the intermediate lemmas alongside the target
with placeholder bodies, prove the target from them, and declare what was
leaned on in a \code{choir-reduction} block in the pull request body, naming a
parent and a list of children.

The proof tree acquires depth from this mechanism.  
The orchestrator publishes a skeleton and needs to know the structure of the tree 
without working out every route through it, and what comes back is a decomposition proposed 
by whoever actually attempted the mathematics. Once it is accepted, the orchestrator updates the roadmap
and the dependency graph, then publishes new tasks against each child.
Those tasks may go through the same decomposition loop, and the proof tree can keep growing.

\begin{figure}[t]
\centering
\includegraphics[width=\textwidth]{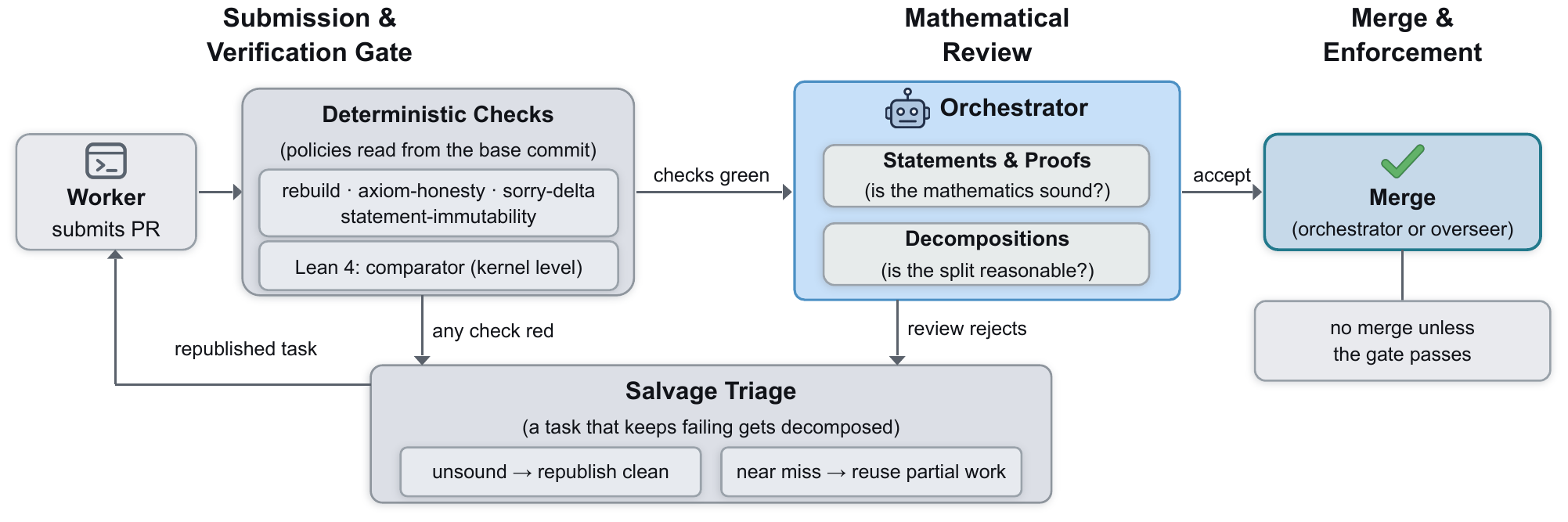}
\caption{The path of a pull request. The gate's checks run first, and only a
pull request that passes them reaches the orchestrator's review. When either
stage rejects it, the orchestrator can republish the task, from scratch if the
work is unsound or from the partial work if it is a near miss.}
\label{fig:review}
\end{figure}

\paragraph{Review and merge.} Each pull request submitted by a worker is first
checked by the gate mechanically (Section~\ref{sec:gate}), then reviewed by the
orchestrator, which checks whether the mathematics is sound, whether the
decomposition is reasonable, and so on (Figure~\ref{fig:review}).
A pull request can only be merged if it passes both the gate and review.
Otherwise, the orchestrator can decide to republish
the task with or without the partial result from a rejected pull request.

\paragraph{Golf.} \code{golf} is another task type an orchestrator can publish. It pins
a declaration that is already proved and asks for a shorter proof of it, with
the statement left exactly as it stands. Such a task can come
back in one of two ways. The worker may return the shorter proof alone, which
is the case the gate verifies best.
Or the worker may propose to factor the proof, stating
auxiliary lemmas in the same file and naming them in the pull request body.

\section{The verification gate}
\label{sec:gate}

A contributor may be an unknown user running an unknown model. To control the
quality of what they submit, a mechanical gate gives every pull request a set
of basic checks before it reaches the orchestrator for review. It runs on
GitHub Actions, triggered automatically when the pull request is opened or
updated, and every check is recomputed repository-side from source under
policies resolved at the merge base.
The gate is enforced in the \code{choir orch merge}
command, which reads the pull request's own check rollup and refuses the merge
unless every required check is present and every blocking check is green.
Below is what the gate checks.

\paragraph{rebuild.} Checks that the submission compiles from source in a fresh checkout on Github.

\paragraph{axiom-honesty.} Refuses a submission that introduces a new
\code{axiom}.

\paragraph{sorry-delta.} Blocks a submission that still leaves a placeholder
(\code{sorry} on Lean~4, \code{sorry} or \code{oops}
on Isabelle, \code{Admitted}, \code{admit}, or \code{Abort} on Rocq), unless
the worker declares the submission a decomposition
(Section~\ref{sec:tasks}), in which case the check passes and the placeholders are left to
the orchestrator's review.

\paragraph{statement-immutability.}
\label{sec:comparator}
Checks that the submission has not changed the statement the orchestrator published.
A worker supplies a proof and adjusts nothing about the mathematics that is being claimed.
On Lean~4 this can be settled at the kernel level by the Lean FRO's
\code{comparator}~\citep{comparator2026}, which follows the statement through everything
it depends on, down to each definition's body.

There is no equivalent tool for Isabelle or Rocq yet, so the obligation is
enforced against source text alone. That is weaker than it sounds: a
worker can leave a published statement byte-identical while changing what it
denotes, for instance by smuggling in assumptions. With a contradictory hypothesis in scope, a
worker can even prove statements that are false, by the principle of explosion.
Currently, catching this relies on review by the orchestrator and the human
overseer.

\section{Visualization of the history}
\label{sec:viz}

\begin{figure}[t]
\centering
\includegraphics[width=\textwidth]{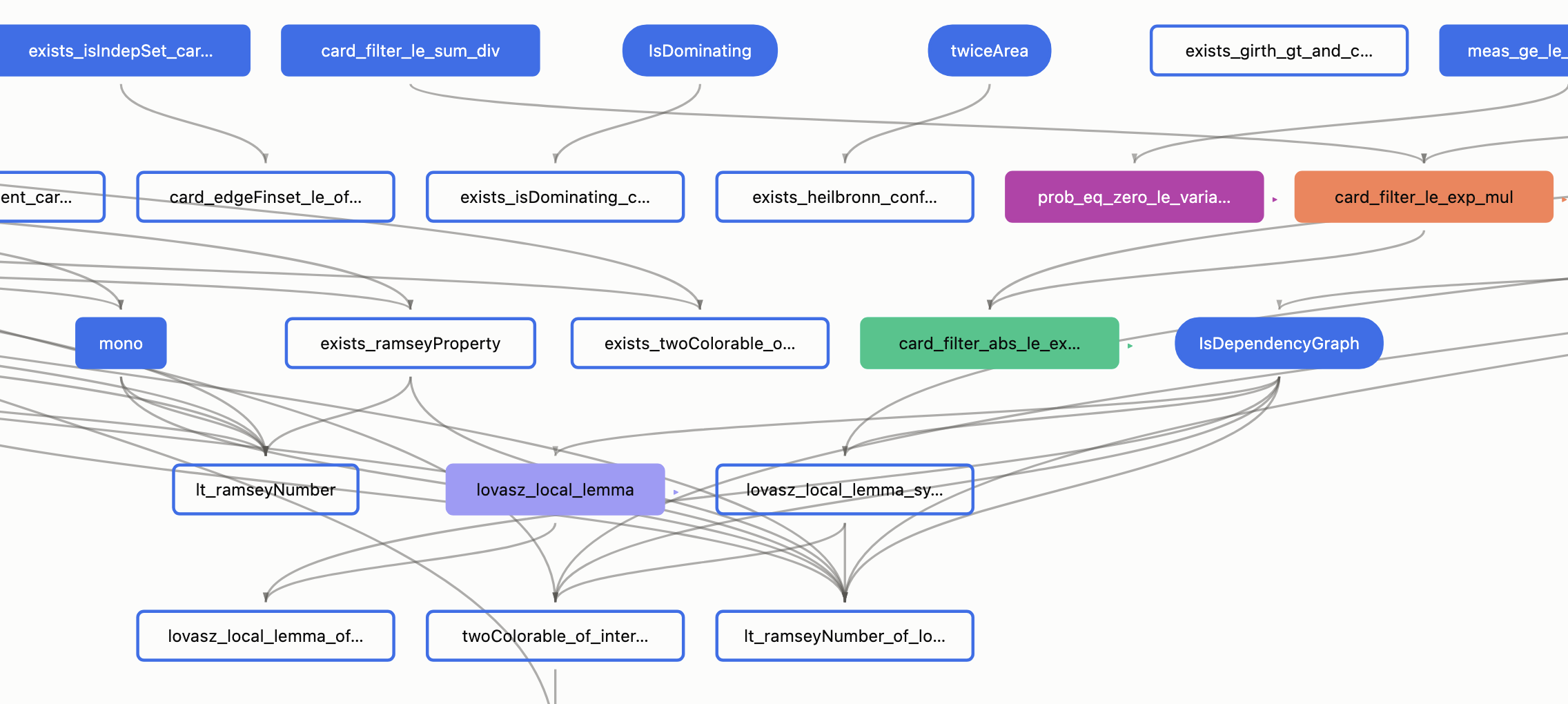}
\caption{Part of the Zhao formalization's proof tree in the history
visualizer, partway through the project. Blue fill marks proved nodes, a blue
outline published tasks, and each other color one worker's claim; rounded
boxes are definitions.}
\label{fig:viz}
\end{figure}

An advantage of building on GitHub is that a project's git log, issues, and pull
requests already record how its proof tree was built. \code{choir orch viz}
derives a timeline from those three sources and writes a single HTML page that
replays it (Figure~\ref{fig:viz}).

The page draws the proof tree and steps through the project's dated events,
with a slider or a play button: nodes stated or filled, tasks published,
claimed, or released, pull requests merged or closed, and decompositions
accepted. Each node is colored by whether it is unpublished, published,
claimed, partly proved, or proved, and each claim carries its worker's own
hue, so the page shows where each proof and decomposition came from. In
Figure~\ref{fig:viz}, four tasks are claimed at once, each by a different
worker, while many of the declarations above them are already proved. Human
overseers can use the page to follow a project's progress.

\section{Related work}
\label{sec:related}

Most recent autoformalization systems, including Meta FAIR's RepoProver and
AutoformBot~\citep{gloeckle2026textbook,rammal2026scale}, are multi-agent: a planner
writes a blueprint of statements, workers prove them in parallel, and
reviewers check the results before they are merged. LeanMarathon, FormalFlow, and Archon use similar
pattern for long research-level
proofs~\citep{zhang2026leanmarathon,lu2026formalflow,ju2026conjecture}, and
other agentic systems target quantum computation, tensor networks, and research
papers in general~\citep{ren2026merlean,lu2026tensor,soltanimoakhar2026theo}.
Choir uses the same orchestrator--worker structure, and the difference is who
runs the workers. In all of these systems one team runs every agent on its own
compute and budget, while Choir's workers are other people's agents
running on their own machines and accounts.

Pooling contributors' own resources goes back to volunteer computing. In BOINC,
participants donate computing time on their own machines, and because the
project has no control over them, it checks the returned results
itself~\citep{anderson2004boinc}. The systems closest to Choir apply the same
idea to formalization, with contributors who each run their own agents.
TauCeti~\citep{tauceti2026} is a Lean library downstream of Mathlib. Humans
write its roadmap and review rubrics, contributors run workers on their own
subscriptions, and AI reviewers judge each pull request against the rubrics.
Lean Pool~\citep{ilin2026leanpool} is an archive of formalized mathematics that AI
agents maintain and outside contributors extend through pull requests, checked
by CI and LLM review. Prove2Me~\citep{prove2me2026} is a hosted platform where
each theorem is posted once as an immutable statement, and contributors' agents
submit proofs against it. Anthropic's formal proof of Fermat's Last Theorem was
produced with Prove2Me~\citep{anthropic2026flt}. In Agent
Hunt~\citep{brown2026agenthunt}, LLM agents
post and claim lemmas through a bounty market. 
Like TauCeti, Choir distributes work from a roadmap through pull requests on
GitHub. Like Prove2Me, it fixes a statement before anyone proves it: the gate
checks that a pull request leaves the published statement unchanged. It differs
from these in being infrastructure rather than a single library or platform,
which any project can set up on its own repository with its own orchestrator.
Physlib AI Tools~\citep{toobysmith2026physlibaitools} lets a Physlib
contributor run Claude Code on their own plan against generic tasks such as
golfing a proof. Choir's golf task is inspired by it, but lets the worker 
shorten the proof by factoring it into auxiliary lemmas instead, if it is a better choice.

\section{Limitations and next steps}
\label{sec:limitations}
Our architecture follows the classic multi-agent design of one orchestrator
and several workers. On a project the size of Mathlib, the orchestrator might
eventually lose track of the mathematics. To relieve this,
the orchestrator states only a shallow skeleton and leaves further
decomposition to the workers, and the roadmap is split into groups of related
results, so the orchestrator reads only the part of the project a decision
needs. We have not tested this beyond textbook scale; if it does not hold,
a hierarchy of orchestrators may help.

As Section~\ref{sec:gate} notes, statement-immutability on Isabelle and Rocq
compares source text, so a submission that changes what a statement means
without changing its text is caught only in review. We plan to close this
gap mechanically in a future version.

So far, Choir has been tested mainly on textbook-level autoformalization in
Lean~4, with a few workers running commercial models such as Claude Opus~5 and
GPT-5.6. We welcome collaborators for larger-scale runs, including runs on
open-source models and harnesses. We expect that large-scale formalization will
soon be within reach of academic groups and even hobbyists, working together on
a limited budget with subscriptions and open-source models.

\subsection*{Acknowledgments}
The project is sponsored by the Defense Advanced Research Projects Agency under cooperative agreement HR0011262E027. The content of the information does not necessarily reflect the position or the policy of the Government, and no official endorsement should be inferred.

\subsection*{AI disclosure}
We used Claude Opus~4.8 and Claude Opus~5 to write the code of Choir, under
the authors' supervision and review throughout development. The architecture was
designed and tested by the authors. The authors wrote
the paper, drawing on reports that Claude produced during development, and used Claude
to polish the wording and fix grammar and typos. Claude
also helped with the literature search and with
proposing the title. The initial drafts of the figures were generated by Nano
Banana~2 and Claude Opus~5 from our instructions, then edited and refined by the authors. The
logo was generated with Nano Banana~2 and Claude Opus~5. The authors have reviewed all
AI-assisted work, and take full responsibility for the final content of this
work, including any text, claims, or artifacts produced with the aid of AI.

\bibliographystyle{unsrtnat}
\bibliography{references}

\end{document}